\pdfoutput=1
\documentclass[11pt]{article}
\usepackage[]{acl}

\usepackage{times}
\usepackage{latexsym}
\usepackage{graphicx}
\usepackage{tabu}
\usepackage{algorithm}
\usepackage{algorithmic}
\usepackage{booktabs}       
\usepackage{amsfonts}       
\usepackage{nicefrac}       
\usepackage{microtype}      
\usepackage{xcolor}         
\usepackage{subfigure}
\usepackage{mathtools}
\usepackage{amsmath}
\usepackage[algo2e]{algorithm2e} 
\usepackage{wrapfig}
\usepackage{graphbox}
\usepackage[normalem]{ulem}
\useunder{\uline}{\ul}{}

\usepackage{url}

\newcommand{\dataset}{{\usefont{T1}{pzc}{m}{n} ART}}

\usepackage{amsmath,amsfonts,bm}

\def\eqref#1{equation~\ref{#1}}

\def\1{\bm{1}}

\def\vc{{\bm{c}}}

\def\ve{{\bm{e}}}

\def\vh{{\bm{h}}}

\def\vy{{\bm{y}}}

\def\mH{{\bm{H}}}

\DeclareMathAlphabet{\mathsfit}{\encodingdefault}{\sfdefault}{m}{sl}
\SetMathAlphabet{\mathsfit}{bold}{\encodingdefault}{\sfdefault}{bx}{n}

\def\gV{{\mathcal{V}}}

\newcommand{\R}{\mathbb{R}}

\usepackage{mathtools}
\usepackage{multirow}
\usepackage{multicol}
\usepackage{booktabs}
\usepackage{subfigure}

\DeclareMathOperator*{\transformerenc}{Trans-Enc}
\DeclareMathOperator*{\transformerdec}{Trans-Dec}

\DeclareMathOperator*{\pool}{Pool}

\usepackage[T1]{fontenc}
\usepackage[utf8]{inputenc}

\usepackage{microtype}

\title{ClarET: Pre-training a Correlation-Aware Context-To-Event Transformer for Event-Centric Generation and Classification}

\author{Yucheng Zhou$^{1}$\thanks{~~Work is done during internship at Microsoft.} ,
        Tao Shen$^{2}$, Xiubo Geng$^{2\dagger}$, Guodong Long$^{1}$, Daxin Jiang$^{2}$\thanks{~~Corresponding author.} \\
         $^{1}$Australian AI Institute, School of CS, FEIT, University of Technology Sydney \\
         $^{2}$Microsoft \\
         {\tt yucheng.zhou.uts@gmail.com, guodong.long@uts.edu.au}\\
         {\tt \{shentao, xigeng, djiang\}@microsoft.com}\\
         }

\begin{document}
\maketitle
\begin{abstract}
Generating new events given context with correlated ones plays a crucial role in many event-centric reasoning tasks. Existing works either limit their scope to specific scenarios or overlook event-level correlations. In this paper, we propose to pre-train a general Correlation-aware context-to-Event Transformer (ClarET) for event-centric reasoning. To achieve this, we propose three novel event-centric objectives, i.e., whole event recovering, contrastive event-correlation encoding and prompt-based event locating, which highlight event-level correlations with effective training. The proposed ClarET is applicable to a wide range of event-centric reasoning scenarios, considering its versatility of (i) event-correlation types (e.g., causal, temporal, contrast), (ii) application formulations (i.e., generation and classification), and (iii) reasoning types (e.g., abductive, counterfactual and ending reasoning). Empirical fine-tuning results, as well as zero- and few-shot learning, on 9 benchmarks (5 generation and 4 classification tasks covering 4 reasoning types with diverse event correlations), verify its effectiveness and generalization ability. 
\end{abstract}

\section{Introduction}

\begin{figure}[t]
    \centering
    \includegraphics[width=1.0\linewidth]{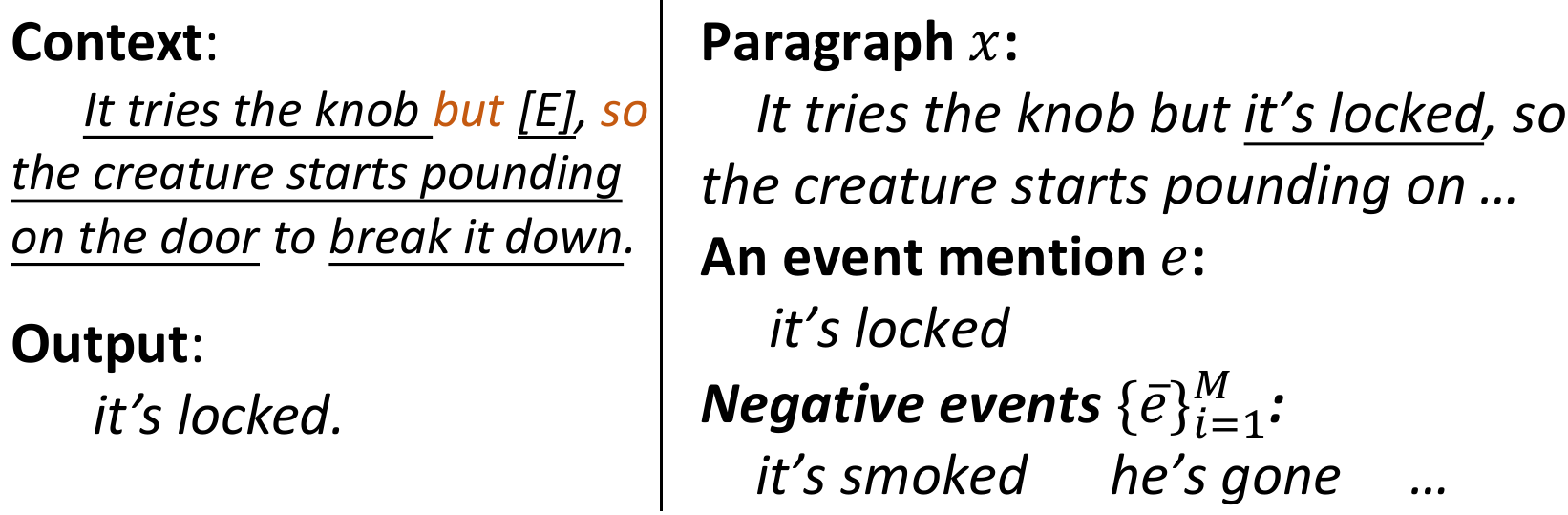}
    \caption{\small \textit{Left}: an example of abductive reasoning which aims to generate the missing event [E] given correlated events (\underline{underlined}) and connectives (w/ \textcolor{orange}{orange}) in the context. 
    \textit{Right}: a toy example $(x,e,\{\bar e\}_{i=1}^M)$ of event-rich data for better reading. See Appendix \ref{app:sec:pre-train-example} for real examples to pre-train our model.}
    \label{fig:intro_case}
\end{figure}

An `event', usually a text span composed of a predicate and its arguments \cite{Zhang20ASER}, is a fine-grained semantic unit to describe the state of entities/things (e.g., \textit{He looks very worried}) and how they act (e.g., \textit{I grab his arms}). Understanding events and modeling their correlations are fundamental to many reasoning tasks \cite{Bhagavatula20Abductive,Qin19Counterfactual}, e.g., abductive reasoning, story ending classification and generation, counterfactual reasoning, script reasoning. For instance, in the left example of Figure~\ref{fig:intro_case}, to generate the missing event \textit{[E]} in the given context, it is essential to understand that there are four events (`\textit{it tries the knob}', \textit{[E]}, `\textit{the creature starts pounding on the door}', and `\textit{(the creature) to break it down}'), and then predict \textit{[E]} based on the other three events and its correlations to them (i.e., the contrast relation indicated by `\textit{but}' and the causal relation by `\textit{so}').

Event-aware reasoning has gained much attention and achieved promising success in recent years \cite{Lv20Integrating,Ding19Event}. However, many algorithms are designed to solve only some specific tasks. For example, \citet{Qin20Back} propose to improve unsupervised decoding for counterfactual and abductive reasoning; \citet{Huang21Story} and \citet{Guan19Story} advance story ending generation via incremental encoding and multi-level graph convolutional networks. Although these works show effectiveness in corresponding applications, they are limited to specific scenarios, and cannot generalize well to a broad scope of reasoning. 

Meanwhile, some pioneering works follow a recently arising paradigm to conduct event-based pre-training for those downstream reasoning tasks \citep{Yu20CoCoLM,han2020deer,Lin2020Conditional,Zhou21EventBERT}. However, these solutions have their own limitations: COMeT \cite{hwang2020comet} learns event correlations from a human-curated knowledge graph and thus limits its scalability. \citet{han2020deer} and \citet{Lin2020Conditional} only model temporal relations and cannot be expanded to other relations (e.g., causal, contrast). EventBERT \cite{Zhou21EventBERT} is proposed for event-based classifications and is thus inapplicable to generation tasks.

In this work, we propose a general pre-training framework for event-centric reasoning by learning a \textbf{C}orre\textbf{la}tion-awa\textbf{r}e context-to-\textbf{E}vent \textbf{T}ransformer (ClarET) from an event-rich text corpus. We propose three novel self-supervised objectives, dubbed as whole event recovering (WER), contrastive event-correlation encoding and prompt-based event locating, respectively. The first one aims to capture event correlation by recovering a whole event from its masked context. The second one enhances the representation of the masked event in WER by contrasting it with the gold event against the negative ones. The last one is a simplified WER task by providing hints in its prompt and thus facilitates effective learning for WER.

ClarET explicitly models event correlations and contributes to various scenarios. From one aspect, it covers a variety of correlation types (e.g., causal, temporal, contrast) attributed to correlation type-agnostic objectives. From another aspect, it is applicable to both generation and classification task formulations by its unified structure. 
Lastly, it highlights event-level correlations and thus is more effective for diverse event-centric tasks, e.g., abductive, counterfactual and ending reasoning.

To evaluate ClarET, we compare it with strong baselines on 9 diverse benchmarks. 
While ClarET is continually pre-trained from BART \citep{Lewis20BART} with very limited extra resources, i.e., training on a small subset of BART-used corpus (i.e., 200M out of 2.2T tokens) within 90 GPU hours (only 0.13\% of 70,000h BART pre-training),
it achieves state-of-the-art (SoTA) performance on all 5 generation benchmarks. It also outperforms all unified models on 4 classification benchmarks and achieves competitive, or even better, accuracy to strong discriminative baselines. 
We further exhibit that the ClarET provides a good initialization for downstream tasks by zero- and few-shot learning.

\section{Related Work}
\paragraph{Unified Pre-trained Model. }

A recent trend is to pre-train unified (a.k.a. universal or general) models to boost downstream generation and classification tasks, rather than masked language modeling (MLM) only.
GPT \cite{radford2019language} is based on auto-regressive language modeling but incompetent in classifications due to unidirectional contextualizing. To remedy this, BART \cite{Lewis20BART} trains seq2seq models as a text denoising autoencoder with mask-infilling, etc; UniLM \cite{Dong19Unified} designs advanced self-attention masks in Transformer, leading to a partially auto-regressive MLM; GLM \cite{Du21All} proposes an auto-regressive blank-filling objective based on Transformer, achieved by bi-/uni-directional attention and 2D positional encoding. T5 \citep{Raffel20Exploring} pre-trains a text-to-text Transformer to recover the masked part of input by decoding.
All these general-purpose pre-trained models focus on relatively short-span masking in random, whereas we focus on masking a whole semantic unit (i.e., event) and propose novel training objectives to circumvent problems in long-span event decoding. 
Besides, they are also vulnerable to pretrain-finetune inconsistency, leading to inferior event-centric performance. 

\paragraph{Task-specific Models for Event Reasoning. }
Many recent works present task-specific neural models for various event-centric reasoning types, including 
(1) abductive reasoning \citep{Ji20Language,Dong21On-the-Fly,Zhu20L2R2},
(2) counterfactual reasoning \citep{Qin19Counterfactual,Qin20Back}, 
(3) ending reasoning \citep{Guan19Story,Wang19T-CVAE,Yao19Plan-and-Write,Huang21Story,Guan20Knowledge,Wang17Integrating,Li18Constructing,Ding19Event,Zhou21Modeling,Chaturvedi17Story,Srinivasan18Simple}, 
(4) incoherence reasoning \citep{mori2020finding}.
However, these methods are designed for the specific reasoning scenarios based on task-specific models so hardly generalize to other scenarios. In contrast, we aim to pre-train a general event-centric model for generalizing to various scenarios.

\paragraph{Event-centric Pre-training. }

With similar scopes, many works focus on event-centric pre-training to promote event-related tasks as `event' is a self-contained semantic unit and also an entry of commonsense reasoning. 
One paradigm is to pre-train on corpora without human-labeling. 
Some methods focus on more specific aspects of events and their correlations. DEER \cite{Han20DEER} performs temporal and event masking predictions for temporal relations. \citet{Lin20Conditional} propose to recover a temporally-disordered or event-missing sequence for temporal and causal relations. \citet{Wang20CLEVE} use AMR structure to design contrastive objectives for the event detection task. However, they are not general enough to various event reasoning tasks. 
In contrast, CoCoLM \cite{Yu20CoCoLM} learns an event-level MLM to generalize more. 
EventBERT \cite{Zhou21EventBERT} states the ineffectiveness of event-level MLM and exploits hard negatives via contrasting, contributing much to downstream multi-choice tasks. 
However, these methods are only competent in discriminative tasks. 
The other paradigm is based on supervised pre-training on similar tasks and then performs knowledge transfer, e.g., COMeT \cite{hwang2020comet}, UnifiedQA \cite{Khashabi20UnifiedQA} and UNICORN \cite{Lourie21UNICORN}, but they require human-curated data. 

\paragraph{Event-rich Corpus.}
Although raw corpora are viewed as off-the-shelf pre-training resources, a key question is how to mine event-rich examples. 
Here, `event-rich' denotes that each example contains various events and entails adequate contexts to support event reasoning via either explicit or implicit event-correlation. 
This is crucial to learning event-correlations and reducing unnecessary overheads. 
Except for human-curated resources (e.g., ATOMIC \cite{sap2019atomic} and ConceptNet \cite{Speer2017ConceptNet}), event-rich corpora are also mined via automatic schemes. 
ASER \cite{Zhang20ASER} builds an event-based graph, where each node is an event extracted from a text and the relation of an event pair is predicted by a PDTB model. 
In contrast, EventBERT \cite{Zhou21EventBERT} operates on pure text so filters out correlation-scarce contexts and extracts verb-rooted events. Besides, it offers event sampling methods for hard negatives.
We adopt this data processing method as both pure-text examples and hard negatives are prerequisites of generic and robust pre-training.

\section{Methodology}

\subsection{Prerequisite: Event-rich Corpus} \label{sec:data_prep}

In this work, we directly adopt event-rich data mining and negative sampling methods from \citet{Zhou21EventBERT} but focus our contributions on enlarging application scope of event-centric tasks and overcoming challenges raised in the new scope.

\paragraph{Event-rich Data Mining.} To mine event-rich data from raw corpus, we employ a story corpus, \textsc{BookCorpus} \cite{Zhu15Aligning}, and take a two-step procedural (i.e., `\textit{filter}' and `\textit{extraction}'). 
It \textit{filters} out correlation-scarce paragraphs according to existence of connectives
(i.e., discourse relation keywords, e.g., \textit{however}, \textit{while}). 
Then, it highlights the event spans in the filtered paragraphs  by \textit{extracting} verb-rooted sub-trees in dependency trees of the paragraphs. 
With a filtered paragraph $x$, we build each example as $(x, e)$ where $e$ is an event mention in $x$.
We obtain 200M tokens (out of 1B in \textsc{BookCorpus}) in 3.9M filtered paragraphs. 
\textbf{\textit{For clear notations}}, we denote a text piece as a lower case letter (e.g., $e$). It is tokenized into a sequence as a bold (e.g., $\ve=[e_1,e_2,\dots]$), where a letter w/ subscript $t$ is the $t$-th token in the sequence. 

\paragraph{Negative Event Sampling. } 

Following \citet{Zhou21EventBERT}, we build a pool of events from the whole corpus and then retrieve negative events by three heuristic schemes. 
Given an event $e$ in $(x,e)$, we sample its negative event, $\bar e$, in light of lexicon-based (20\% time), PoS-based (60\% time) or in-domain (20\% time) retrieval. 
Consequently, given an event $e$, we sample $M$ negative events, i.e., $\{\bar e\}_{i=1}^M$. 
Figure~\ref{fig:intro_case} (right) shows an integrated instance $(x,e,\{\bar e\}_{i=1}^M)$ of the event-rich corpus\footnote{Experimental codes are released at \url{https://github.com/yczhou001/ClarET}.
}.

\subsection{Pre-training Objectives}\label{sec:pretrain_obj}

\begin{figure}[t]
    \centering
    \includegraphics[width=1\linewidth]{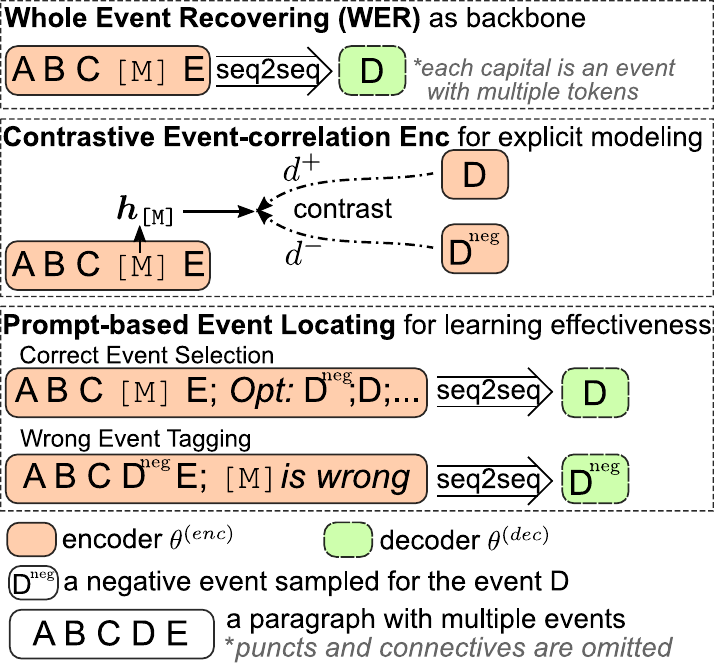}
    \caption{\small An overview of self-supervised objectives for our \textbf{C}orre\textbf{la}tion-awa\textbf{r}e context-to-\textbf{E}vent \textbf{T}ransformer (ClarET). 
    }
    \label{fig:model_overview}
\end{figure}

We first present \textit{whole event recovering} as a backbone pre-training objective in \S\ref{sec:whole_event_recover}. 
After identifying incompetence of the simple backbone, we propose two other objectives in \S\ref{sec:correlate_enc_enhance} and \S\ref{sec:promp_event_locate}. 
An overview of the objectives is shown in Figure~\ref{fig:model_overview}. 

\subsubsection{Whole Event Recovering} \label{sec:whole_event_recover}

For the objective of whole event recovering (WER), it is straightforward to leverage an encoder-decoder structure, where a masked context is passed into the encoder to generate the missing part by decoding. 
Specifically, given an event $e$ in a paragraph $x$, we mask out $e$ from $x$ at the encoder side and then generate $e$ at the decoder side, i.e., 
\begin{align}
    p(e|x_{/\{e\}};\theta) = \prod\nolimits_t p(e_t|e_{<t},x_{/\{e\}};\theta), \label{equ:obj_cag_base}
\end{align}
where $\theta$ denotes parameters and $x_{/\{e\}}$ denotes replacing $e$ in $x$ with \textit{one} special token \texttt{[M]}. 
We estimate Eq.~(\ref{equ:obj_cag_base}) by the Transformer sequence-to-sequence (seq2seq) structure \citep{Vaswani17Attention}.
First, we apply the Transformer encoder to $x_{/\{m\}}$ for contextual embeddings for all tokens in $x_{/\{m\}}$:
\begin{align}
    \mH^{(enc)} \!\!=\!\! \transformerenc(x_{/\{e\}};\theta^{(enc)})\in\R^{d\times n}, \label{equ:trans_enc}
\end{align}
where $n$ is the number of tokens in $x_{/\{e\}}$. Then, the Transformer decoder is employed to predict all tokens $\ve$ of the event $e$ in a recurrent manner, i.e.,
\begin{align}
    \tilde \vy_t = \transformerdec(e_{<t},\mH^{(enc)};\theta^{(dec)})\!\in\!\!\R^{|\gV|},
\end{align}
where $\gV$ denotes token vocabulary and $\tilde \vy_t$ is the predicted categorical distribution over $\gV$. 
Lastly, the training objective is defined as a maximum likelihood estimation. Its loss function is written as
\begin{align}
    L^{(wer)} = - \sum_{(x,e)} \dfrac{1}{|\ve|}\sum\nolimits_{t=1}^{|\ve|}\log \tilde\vy_t[y=e_t], \label{equ:loss_recover}
\end{align}
where `$\vy_t[y=e_t]$' denotes fetching the probability of the $t$-step gold token $e_t\in\ve$ from $\tilde \vy_t$.

This objective is similar to span recovering schema \citep{Raffel20Exploring,Joshi2020spanbert} but differs in that 
(i) each masked span is an event, i.e., an integrated semantic unit, so much longer (up to 22 tokens and see Figure~\ref{fig:long_span_gen} for length distribution), 
and (ii) only one event is masked out from the context to facilitate event-correlation modeling between the event and its contexts.

Intuitively, the success of Eq.~(\ref{equ:obj_cag_base}) requires to capture correlations between the masked event and remaining contexts but two major problems arise due to WER with long event-level masking spans:

\textit{(1) Implicit Event-correlation:} 
The model recovers an event based solely on token-level concurrence as in a conditional language model (e.g., T5 and BART), regardless of the rich event-level correlations between the events in context $x_{/\{e\}}$ and the masked event $e$. 
Such a correlation-implicit model would achieve inferior performance on downstream event-centric correlation reasoning tasks. 

\textit{(2) Learning Difficulty:} 
As the masked event is an integrated, self-contained, semantic unit, it is difficult for the conditional generation model to recover the whole event due to a lack of local contexts. As a result, the model cannot effectively learn from the long masked spans,
which has been empirically proved in autoencoding MLM models. 

To alleviate the two problems above, we propose two other novel self-supervised objectives in the following. 
Briefly, we present contrastive event-correlation encoding to enhance correlations between contexts and events, and prompt-based event locating to reduce generation difficulty.

\subsubsection{Contrastive Event-correlation Encoding} \label{sec:correlate_enc_enhance}

For the \textit{implicit event-correlation} problem, an intuitive solution is to explicitly highlight the correlation from the masked context to the missing event at the encoder side. 
To achieve this, we resort to contrastive learning to enhance the encoder-side representation of the masked event by contrasting it with the embedding of the gold event mention $e$ against those of negative ones $\bar e$.
Particularly, we first derive the embedding of $e$ and $\bar e$ independently via the Transformer encoder in Eq.(\ref{equ:trans_enc}), i.e.,
\begin{align}
 \vc =\pool(\transformerenc(\texttt{[CLS]}+e;\theta^{(enc)})),  \\
 \bar \vc =\pool(\transformerenc(\texttt{[CLS]}+\bar e;\theta^{(enc)})), 
\end{align}
where \texttt{[CLS]} is a special token prefixed to each event mention, and $\pool(\cdot)$ denotes using the contextual embedding of \texttt{[CLS]} to represent the whole event. 
Then, we enhance $\vh_{\texttt{[m]}}$, the contextual representation of \texttt{[M]} in $x_{/\{e\}}$ from $\mH^{(enc)}$ in Eq.(\ref{equ:trans_enc}), by contrasting it with $\vc$ against $\bar \vc$, i.e.,
\begin{align}
    L^{(cee)} \!\!=\!\! \max (0,\!\lambda \!+\! d(\vh_{\texttt{[m]}}, \!\vc) \!-\! d(\vh_{\texttt{[m]}}, \!\bar\vc)),  \label{equ:loss_contrast} 
\end{align}
where $d(\cdot,\cdot)$ denotes a distance metric of two vectors, which is Euclidean distance in this work. As a result, the encoder-side correlation-aware representation $\vh_{\texttt{[m]}}$ also offers a straightforward pathway to transmit event-level information to decoding so mitigates the \textit{learning difficulty} to some extent.

\subsubsection{Prompt-based Event Locating}  \label{sec:promp_event_locate}
As for \textit{learning difficulty} problem, we also propose a prompt-based event locating objective to reduce generative difficulty by providing hints in the prompt. 
The basic idea is to simplify WER objective as an extractive generation task to locate and copy a candidate/hint from the prompt, which aims at improving learning effectiveness.
To this end, we present two prompt-based generation schemas in the following. 

\paragraph{Correct Event Selection.} 
Inspired by advances of prompt-based multi-choice question answering, we present correct event selection schema to select the gold event $e$ against negative ones $\{\bar e\}_{i=1}^M$ based on the contexts $x_{/\{e\}}$. 
Given an event-masked paragraph $x_{/\{e\}}$ suffixed with several candidate events $\{\bar e\}_{i=1}^M$ containing the gold masked one $e$, it aims to generate the masked event $e$ back, i.e.,
\begin{align}
    \notag \hat x^{(ces)} = x_{/\{e\}} + \underline{\text{Options:~} \text{(a)~} e^{1} \text{;~(b)~} e^{2}\text{;~}\cdots},
\end{align}
where $[e^{1}, e^{2}, \dots]$ is a random permutation of $[e,\{\bar e\}_{i=1}^M]$ in case of position bias. We use a random permutation as all candidates are assigned with distinct position embeddings during contextualizing, and a fixed permutation of gold events will result in a learning shortcut (position bias) to degrade the model. Thus, similar to Eq.(\ref{equ:obj_cag_base}), we can define its formula as $p(e|\hat x^{(ces)};\theta)$.

\begin{table*}[t] \small
\centering
\setlength\tabcolsep{2pt}
\begin{tabular}{lccccccccccccc}
\toprule
\textbf{}   & \multicolumn{1}{l}{} & \multicolumn{3}{c}{\textbf{\begin{tabular}[c]{@{}c@{}}Abductive C.S.\\ Reasoning\end{tabular}}} & \multicolumn{3}{c}{\textbf{\begin{tabular}[c]{@{}c@{}}Counterfactual\\ Story\end{tabular}}}   & \multicolumn{2}{c}{\textbf{\begin{tabular}[c]{@{}c@{}}Story Ending\\ Generation\end{tabular}}} & \multicolumn{2}{c}{\textbf{\begin{tabular}[c]{@{}c@{}}C.S. Story\\ Generation\end{tabular}}} & \multicolumn{2}{c}{\textbf{\begin{tabular}[c]{@{}c@{}}Event Process\\ Completion\end{tabular}}}  \\\cmidrule(lr){3-5} \cmidrule(lr){6-8} \cmidrule(lr){9-10} \cmidrule(lr){11-12} \cmidrule(lr){13-14} 
           & \multicolumn{1}{l}{\textbf{Size}}  & \textbf{B-4}                     & \textbf{R-L}                       & \textbf{BERT}                    & \textbf{B-4}    & \textbf{R-L}     & \textbf{BERT}  & \textbf{B-1}                                   & \textbf{B-2}                                  & \textbf{B-1}                                     & \textbf{B-2}                                     & \textbf{B-1}                                   & \textbf{B-2}                                   \\\midrule\midrule
\multicolumn{14}{l}{\textit{Selected task-specific models with competitive performance}}  \\\midrule
GRF \cite{Ji20Language}      & -   & 11.62               & 34.62                & -                 & -               & -                & -              & -                          & -                         & -                            & -                            & -                          & -                          \\
IE+MSA \cite{Guan19Story}   & -   & -                   & -                    & -                 & -               & -                & -              & 24.40                      & 7.80                      & -                            & -                            & -                          & -                          \\
Plan\&Write  \cite{Yao19Plan-and-Write} & - & -                   & -                    & -                 & -               & -                & -              & 24.40                      & 8.40                      & 30.80                        & 12.60                        & -                          & -                          \\\midrule\midrule
\multicolumn{14}{l}{\textit{Fine-tuning with pre-trained unified (generative) model}}  \\\midrule
GPT2-S \cite{radford2019language}   & 124M   & 2.23                & 22.83                & 48.74             & 69.27           & 65.72            & 60.53          & 39.23                      & 13.08                     & 32.20                        & 14.10                        & 35.25                      & 11.75                      \\
GPT2-M \cite{radford2019language} & 335M     & -                   & -                    & -                 & 75.71           & 72.72            & 62.39          & -                          & -                         & -                            & -                            & 45.43                      & 14.81                      \\
BART \cite{Lewis20BART} & 400M & {\ul 16.47}               & {\ul 38.73}                & {\ul 56.36}             & {\ul 82.91}           & {\ul 76.44}            & {\ul 79.50}          & 54.22                      & 18.07                     & 54.22                        & 18.07                        & 56.25                      & 18.75                      \\
GLM \cite{Du21All} & 335M  & 7.79              & 25.54                & 54.85             & 75.81           & 70.03            & 68.23          & {\ul 57.04}                      & {\ul 18.45}                     & {\ul 57.04}                        & {\ul 18.45}                        & {\ul 57.34}                      & {\ul 19.11}                      \\
\midrule
\textbf{ClarET} (ours) & 400M  & \textbf{17.67}      & \textbf{41.04}       & \textbf{57.31}    & \textbf{87.18}  & \textbf{80.74}   & \textbf{81.48} & \textbf{57.47}             & \textbf{19.16}            & \textbf{57.47}               & \textbf{19.16}               & \textbf{58.88}             & \textbf{19.74}             \\
\bottomrule
\end{tabular}
\caption{\small Fine-tuning results on five generation benchmark datasets. Previous state-of-the-art (SoTA) results are \underline{underlined}, `Size' denotes the number of model parameters, and `C.S.' is an abbreviation of CommonSense. Please refer to Appendix \ref{app:sec:gen_task} for the reported results of more task-specific models on each dataset.
}
\label{tab:gen_result}
\end{table*}

\paragraph{Wrong Event Tagging.} 
The other schema is wrong event tagging to find the wrong event in a corrupted paragraph, similar to incoherence reasoning. 
Thus, we re-write the encoder input as
\begin{align}
    \notag \hat x^{(wet)} = x_{/\{e\} \& \cup \{\bar e\}} + \underline{\text{Event:~}\texttt{[M]}\text{~is wrong}},
\end{align}
where $x_{/\{e\} \& \cup \{\bar e\}}$ denotes replacing the gold event $e$ in $x$ with a negative $\bar e \in \{\bar e\}_{i=1}^M$. Thus, we can define the formula of this objective as $p(\bar e|\hat x^{(wet)};\theta)$. 

Based on the two formulas above, we define the prompt-based event locating objective as
\begin{align} 
    \notag L^{(pel)} &= \sum_{(x,e)} - \dfrac{1}{|\ve|} \sum_t \log p(e_t|e_{<t},\hat x^{(ces)};\theta)  \\
    &- \dfrac{1}{|\bar\ve|} \sum_t \log p(\bar e_t|\bar e_{<t},\hat x^{(wet)};\theta), \label{equ:loss_prompt}
\end{align}
where $\theta = \{\theta^{(enc)}, \theta^{(dec)}\}$, $\bar e$ is sampled in $\{\bar e\}_{i=1}^M$.

\subsection{Model Pre-training and Fine-tuning} \label{sec:pretrain_infer}

\paragraph{Self-supervised Pre-training.}
The final loss to pre-train our ClarET is a linear combination of the three losses above from Eq.(\ref{equ:loss_recover}, \ref{equ:loss_contrast}, \ref{equ:loss_prompt}), i.e., 
\begin{align}
    L = L^{(wer)} + L^{(cee)}  + L^{(pel)}. \label{equ_loss_all}
\end{align}
We set the margin $\lambda$ in Eq.(\ref{equ:loss_contrast}) to 0.5 w/o tuning. 

\paragraph{Supervised Downstream Fine-tuning. }
For generation tasks, we simply leverage the formula in Eq.(\ref{equ:obj_cag_base}) to establish fine-tuning objectives. 
For discriminative (e.g., multi-choice) tasks, we can either formulate all tasks into generation as in GPT/T5 or fine-tune with classifying heads as in BART. With pilot experiments, we found the latter one can achieve better performance and adopted it.

\subsection{Comparing to Similar Works} \label{sec:detail_comp}

While we adopt the same data processing in EventBERT \citep{Zhou21EventBERT} and share a similar motivation to learn an event-centric pre-trained model, 
we expand the scope from `\textit{discriminative-only}' in EventBERT into `\textit{unified}' by our context-to-event Transformer for a broad spectrum of scenarios. 
Such an expansion is non-trivial since new challenges arise in the unified formulation.
Compared to the inefficient `\textit{event-backfilling and contextualizing}' paradigm in EventBERT, our model can explicitly and effectively learn event-level correlations between contexts and events by our novel contrastive and prompt-based objectives. 
Moreover, COMeT \citep{bosselut2019comet,hwang2020comet} is also a conditional generation model but focuses on triple-level commonsense reasoning --  given (\textit{head event}, \textit{relation}) to generate \textit{tail events}, whose motivation, however, is orthogonal to ours. 
Therefore, we focus on a different motivation or scope, not to mention evaluation formulations. 

\section{Experiments}

This section begins with descriptions of downstream datasets and experimental setups.

\begin{table*}[t]\small
\centering
\setlength\tabcolsep{2pt}
\begin{tabular}{lccccc}
\toprule
& &  \textbf{\begin{tabular}[c]{@{}c@{}}Abductive C.S. \\ Reasoning\end{tabular}} & \textbf{\begin{tabular}[c]{@{}c@{}}Script \\ Reasoning\end{tabular}} & \textbf{\begin{tabular}[c]{@{}c@{}}Narrative Incoherence \\ Detection\end{tabular}} & \textbf{\begin{tabular}[c]{@{}c@{}}Story Cloze \\ Test\end{tabular}} \\ \cmidrule(lr){3-3} \cmidrule(lr){4-4} \cmidrule(lr){5-5} \cmidrule(lr){6-6}
& \textbf{Size} & \textbf{ACC (\%)}                                                                 & \textbf{ACC (\%)}                                                         & \textbf{ACC (\%)}                                                                        & \textbf{ACC (\%)}                                                         \\\midrule\midrule
\multicolumn{5}{l}{\textit{Selected task-specific models with competitive performance}}                                                                                                                                                                                                                                                                        \\\midrule
Hidden   Coherence Model    \cite{Chaturvedi17Story}   & -         & -                                                                            & -                                                                    & -                                                                                   & 77.60                                                                \\
GRU   Context \cite{mori2020finding}     & -                     & -                                                                            & -                                                                    & 52.20                                                                               & -                                                                    \\
RoBERTa + Kown. Model \cite{Zhou21Modeling}     & 469M       & -                                                                            & {\uwave{63.62}}                                                                & -                                                                                   & -                                                                    \\\midrule\midrule
\multicolumn{5}{l}{\textit{Fine-tuning with pre-trained discriminative model}}                                                                                                                                                                                                                                                                               \\\midrule
RoBERTa \cite{Liu19RoBERTa}     & 345M                     & 82.35                                                                        & 61.53                                                                & 73.94                                                                               & 87.10                                                                \\
EventBERT \cite{Zhou21EventBERT}    & 345M                          & {\uwave{85.51}}                                                               & 63.50                                                                & {\uwave{75.03}}                                                                   & {\uwave{91.33}}                                                       \\\midrule\midrule
\multicolumn{5}{l}{\textit{Fine-tuning with pre-trained unified model}}                                                                                                                                                                                                                                                                               \\\midrule
CALM \cite{Zhou21Pre-training}   & 770M                          & 77.12                                                                        & -                                                                    & -                                                                                   & -                                                                    \\
UNICORN \cite{Lourie21UNICORN}    & 770M                            & 79.50                                                                        & -                                                                    & -                                                                                   & -                                                                    \\
BART \cite{Lewis20BART}   & 400M                           & {\ul{80.74}}                                                                        & {\ul{61.34}}                                                                & {\ul{72.48}}                                                                               & {\ul{87.01}}                                                                \\\midrule
\textbf{ClarET} (ours)  & 400M              & \textbf{82.77}                                                                        & \textbf{64.61}                                                       & \textbf{74.88}                                                                               & \textbf{91.18}                                                                \\\bottomrule
\end{tabular}
\caption{\small Fine-tuning results on four classification benchmark datasets. 
We split pre-trained models into discriminative and unified groups since discriminative models usually outperforms unified ones in classification and our ClarET falls into the latter. 
Previous SoTA discriminative and unified results are \uwave{waved} and \underline{underlined}, respectively.
See Appendix \ref{app:sec:cls_task} for full results.
}
\label{tab:cls_result}
\end{table*}

\paragraph{Downstream Datasets.}
We conduct extensive evaluations on 9 datasets for 9 downstream tasks, i.e., 5 generation and 4 classification tasks.  
Generation tasks include abductive commonsense reasoning on \emph{\dataset{}} ($\alpha$NLG) \cite{Bhagavatula20Abductive}, 
counterfactual story generation on TIMETRAVEL \cite{Qin19Counterfactual}, 
story ending generation \cite{Guan19Story}, 
commonsense story generation \cite{Guan20Knowledge}, 
and event process completion on APSI \cite{Zhang20Analogous}. 
Classification tasks include script reasoning on MCNC \cite{Li18Constructing}, 
abductive commonsense reasoning on \emph{\dataset{}} ($\alpha$NLI) \cite{Bhagavatula20Abductive}, 
narrative incoherence detection on ROCStories \cite{mori2020finding}, 
and story cloze test \cite{Mostafazadeh16Corpus}. Please refer to Appendix \ref{app:sec:dataset} for their details.

\paragraph{Pre-training Setups.}
Instead of learning from scratch, we perform continual pre-training from BART-large \citep{Lewis20BART} due to limited computation resources. 
The batch size and number of training steps are 1152 and 160k.
The model is trained by Adam \cite{Kingma14Adam} w/ learning rate of 1e-5 and warmup proportion of 0.03.
The gradient clip, dropout rate and weight decay are 1.0, 0.1 and 0.01.
Notably, (i) \textsc{BookCorpus} has already been used by BART pre-training and our data processing is based on heuristics without human-curated resources;
(ii) Our continual pre-training only needs 90 GPU hours on 200M tokens, i.e., 0.13\% of BART that consumes 70K hours on 2.2T tokens (see Appendix~\ref{app:sec:bart_pretrain_resource}).
Hence, ClarET with zero newly introduced corpus and relatively negligible computing overhead makes great lifts and preserves fair comparisons with baselines.

\paragraph{Fine-tuning Setups.}
For finetuning, we train the model with an Adam w/ learning rate of 1e-5 and warmup proportion of 0.06. 
The dropout rate, batch size and weight decay are 0.1, 32 and 0.01.
For generative downstream tasks, we take BLEU-$N$ (B-$N$) \cite{Papineni02Bleu}, ROUGE-L (R-L) \cite{lin04rouge} and BERTScore (BERT) \cite{Zhang20BERTScore} as the evaluation metrics, while the accuracy (ACC) is taken for classification tasks. 
Each fine-tuning runs with seeds 2, 10 and 1234, and we evaluate the best dev model on the test set.

\begin{table}[t]\small
\centering
\setlength\tabcolsep{2pt}
\begin{tabular}{lrrr}
\toprule
\textbf{Method}  & \textbf{B-4} & \textbf{R-L} & \textbf{BERT} \\
\midrule
GPT \cite{Qin19Counterfactual}             & 1.25            & 18.26            & 59.50         \\
GPT2-S \cite{Qin19Counterfactual}           & 1.28            & 20.27            & 59.62         \\
GPT2-M \cite{Qin19Counterfactual}          & 1.51            & 19.41            & 60.17         \\
Zero-Shot-Ranked \cite{Qin20Back} & 2.26            & 25.81            & 60.07         \\
BART-large \cite{Lewis20BART}      & 7.08            & 30.60            & 61.58         \\
\textsc{DeLorean} \cite{Qin20Back}        & 21.35           & 40.73            & 63.36         \\\midrule
\textbf{ClarET} (ours)        & \textbf{23.75}           & \textbf{43.03}            & \textbf{63.93}         \\\bottomrule
\end{tabular}
\caption{\small Zero-shot results on generative Counterfactual Story.}
\label{tab:zero_cs}
\end{table}

\subsection{Main Evaluation}

\paragraph{Fine-tuning for Generation.}
As shown in Table~\ref{tab:gen_result}, our proposed ClarET achieves SoTA performance across all generation tasks. For instance, ClarET increases the ROUGE-L score by 2.3 absolute value for abductive reasoning. 
The superior performance of ClarET on the benchmarks demonstrates that it can model event-level correlation more effectively via few steps of continual pre-training and provide a general solution for a variety of event-centric correlation reasoning tasks.

\paragraph{Fine-tuning for Classification.}
Table~\ref{tab:cls_result} lists results on 4 classification tasks. We find ClarET performs better than all task-specific models and unified pre-trained models with 2\%-4\% improvement. It achieves competitive accuracy to strong discriminative models, e.g., the gap between ClarET and EventBERT is  $\sim$0.15 for narrative incoherence detection and story cloze test. 
However, EventBERT is a RoBERTa-based competitor using the identical pre-training corpus. Its pre-training follows ``event-backfilling and contextualizing'' (similar to multi-choice QA), which has a small gap to downstream classification tasks for strong performance but brings two drawbacks. 
Firstly, its pre-training is slow due to repeat contextualizing over paragraphs, leading to $5.6\times$ longer GPU hours than ours.
In addition, its discriminative paradigm limits it specifically to classifications, regardless of wide generation tasks.
The results show ClarET is on par with the discriminative-only EventBERT on classifications. This is non-trivial given the large formulation gap between our generative pre-training objectives and downstream multi-choice-style classification tasks, and attributed to our effective event-correlation learning. 
In summary, these results show ClarET serves as a unified pre-trained model for event-centric generation and classification tasks.

\subsection{Quantitative Analysis}

\begin{table}[t]\small
\centering
\begin{tabular}{lc}
\toprule
\textbf{Method} & \textbf{ACC (\%)} \\
\midrule
Random          & 20.00             \\
RoBERTa-large \cite{Zhou21EventBERT}  & 20.09             \\
DeBERTa-xlarge \cite{Zhou21EventBERT} & 20.31             \\
BART-large \cite{Lewis20BART}     & 21.72             \\
EventBERT \cite{Zhou21EventBERT}      & 30.79             \\\midrule
\textbf{ClarET} (ours)       & \textbf{32.15}    \\\bottomrule
\end{tabular}
\caption{\small Zero-shot results on discriminative Script Reasoning.  Note that MLM-style models are evaluated by autoregression-like operation \citep{Zhou21EventBERT}.}
\label{tab:zero_sr}
\end{table}

\paragraph{Zero-shot Learning.}
It is essential to verify if the targeted information was learned and retained by a pre-trained model. Compared to MLM, our generative recovering model is inherently applicable to event-centric multi-choice and generative formulations. 
\textit{For generation tasks}, we apply Eq.(\ref{equ:obj_cag_base}) to generate answers. As shown in Table~\ref{tab:zero_cs}, ClarET achieves the best performance and outperforms \textsc{DeLorean} (which adapts auto-regression for counterfactual reasoning). 
\textit{For classification tasks}, we apply Eq.(\ref{equ:obj_cag_base}) to each option for its perplexity and select the option with minimum. 
As shown in Table~\ref{tab:zero_sr}, ClarET surpasses previous models and beats the discriminative-only event-centric model, EventBERT. Besides, the general-purpose pre-trained models perform nearly random guesses due to their incompetence in long-span event discrimination. 

\begin{figure}[t]
\centering
\centering
\includegraphics[width=0.48\linewidth]{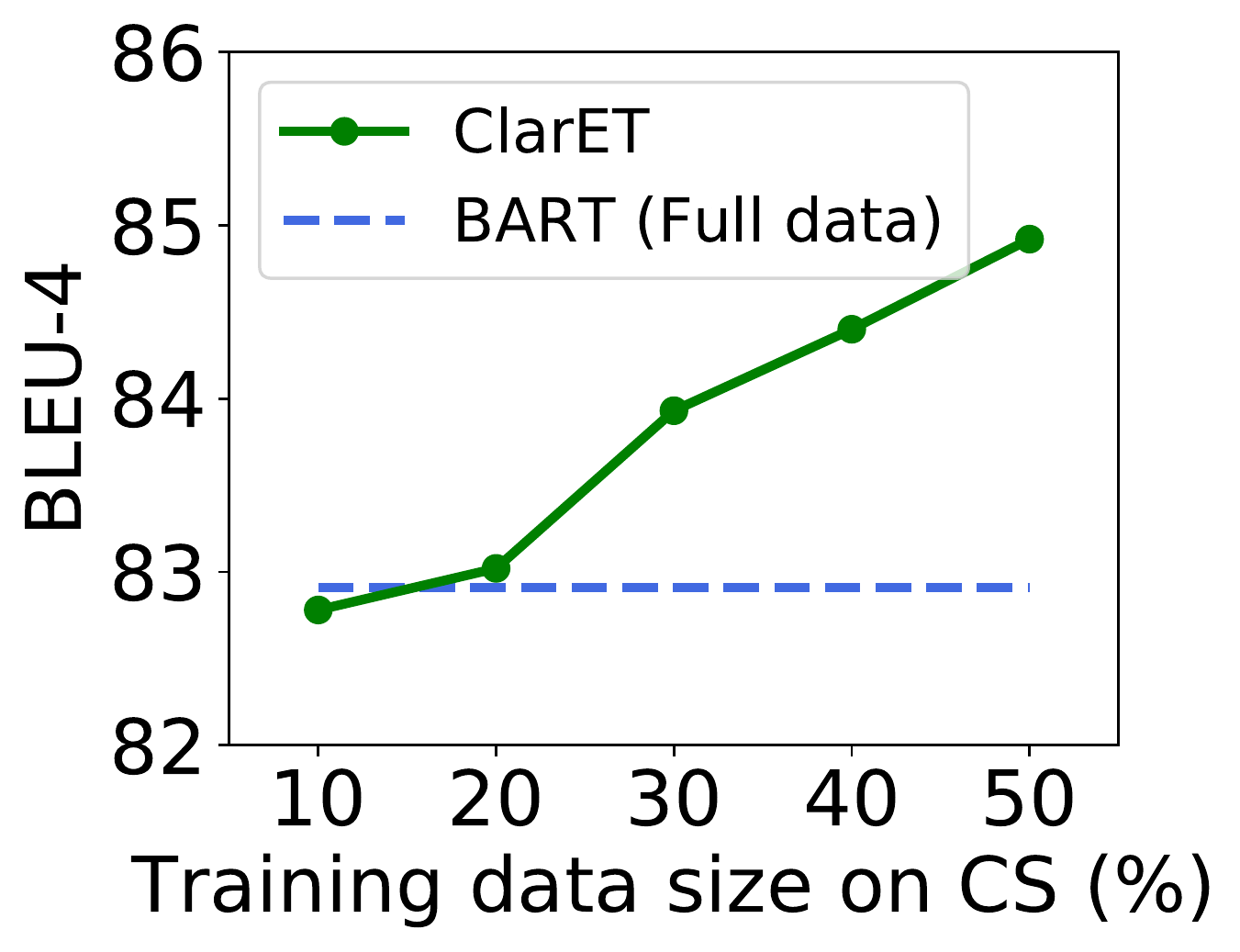}
\includegraphics[width=0.48\linewidth]{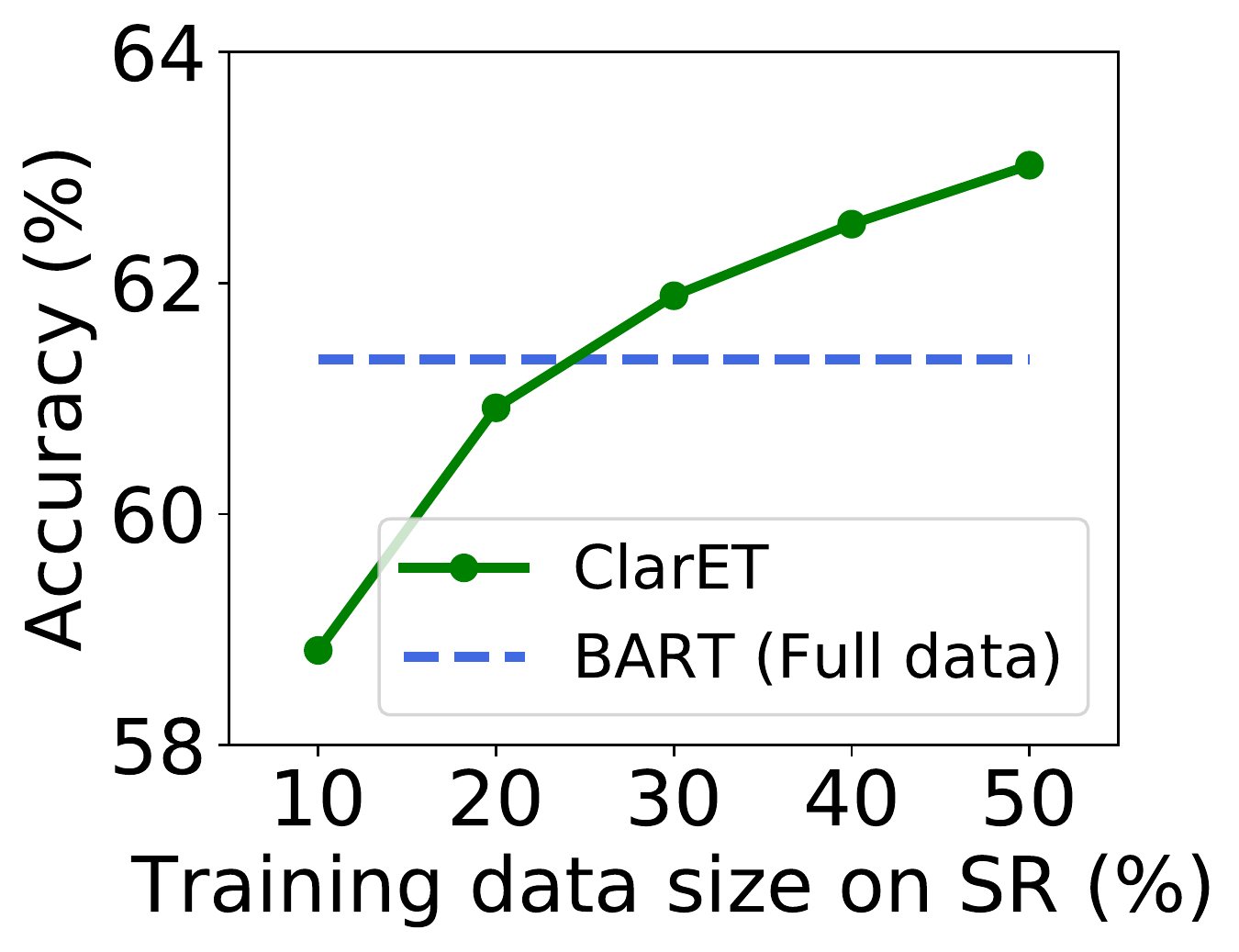}
\caption{\small Few-shot learning results compared with the basic model, BART-large, on generation (Counterfactual Story, CS) and classification (Script Reasoning, SR).}
\label{fig:few_shot_result}
\end{figure}
\paragraph{Few-shot Learning.}
Since our model reduces pretrain-finetune inconsistency for event-centric tasks and provides a good initialization for downstream fine-tuning, it is also interesting to see few-shot performance by scaling down training data. 
As shown in Figure~\ref{fig:few_shot_result}, ClarET achieves similar performance to strong baselines with only 10\%-30\% of training data for fine-tuning. 

\begin{table}[t]\small
\centering
\setlength\tabcolsep{2pt}
\begin{tabular}{lccc}
\toprule
\multicolumn{1}{c}{\multirow{2}{*}{\textbf{Method}}} & \multicolumn{2}{c}{\textbf{Gen-CS}} & \textbf{Cls-SR} \\\cmidrule(lr){2-3} \cmidrule(lr){4-4}
\multicolumn{1}{c}{}                                 & \textbf{B-4}       & \textbf{R-L}       & \textbf{ACC}       \\\midrule
ClarET (full, pre-trained by Eq.(\ref{equ_loss_all}).)                                & \textbf{87.18}     & \textbf{80.74}     & \textbf{64.61}          \\\midrule
$\Diamond$ w/o correct event selection (prompt)                 & 86.76              & 80.03              & 63.06                   \\
$\Diamond$ w/o wrong event tagging (prompt)                     & 86.33              & 79.84              & 63.89                   \\
$\Diamond$ w/o contrastive encoding                             & 85.84              & 78.69              & 63.24                   \\
$\Diamond$ only prompt-based event locating                     & 83.32              & 76.51              & 62.97                   \\\midrule
BART-large (basic model)                                          & 82.91              & 76.44              & 61.34                   \\\bottomrule
\end{tabular}
\caption{\small Ablation study of the pre-training objectives in ClarET, which is evaluated by fine-tuning on generation (Counterfactual Story, CS) and classification (Script Reasoning, SR).}
\label{tab:ablation_study}
\end{table}

\paragraph{Ablation study. }
To measure the contribution of each objective to the final fine-tuning results, we conduct an ablation study on both generation and classification in Table~\ref{tab:ablation_study}. 
The first two ablations drop the two prompt schemas respectively in prompt-based event locating objective of Eq.(\ref{equ:loss_prompt}), which verifies the effectiveness of reducing task difficulty. 
Then, the third ablation removes contrastive event-correlation encoding and shows a substantial drop, which verifies the significance of explicit event-correlation learning. 
Next, we keep only the prompt-based event locating objective to make our model a prompt-learning discriminative model (sharing more close methodology with EventBERT), however leading to a dramatic decrease. 
Lastly, when removing all the objectives, our model degenerates to BART-large.

\begin{table}[t] \small
\centering
\begin{tabular}{lc}
\toprule
\textbf{Method} & \textbf{ePPL on Dev} \\
\midrule
\textbf{ClarET} (full model)                             & 8.27               \\
WER-Only Model                            & 8.76               \\\bottomrule
\end{tabular}
  \caption{\small Event generation of ClarET and whole event recovering (WER-only) model on a pre-training event-masked dev set (2\% held-out masked paragraphs by following \citet{Zhou21EventBERT}). The `ePPL', i.e., event perplexity, refers to event-level token perplexity averaged over the dataset. 
}
\label{tab:Event_ppl}
\end{table}

\begin{table}[t]\small
\centering
\setlength\tabcolsep{4pt}
\begin{tabular}{lcccccc}
\toprule
\multirow{2}{*}{} &               & \textbf{ACR}   & \textbf{CS}    & \textbf{SEG}   & \textbf{CSG}   & \textbf{EPC}   \\ \cmidrule(lr){3-3} \cmidrule(lr){4-4} \cmidrule(lr){5-5} \cmidrule(lr){6-6} \cmidrule(lr){7-7}
                                 & \textbf{Size} & \textbf{R-L}   & \textbf{B-4}   & \textbf{B-1}   & \textbf{B-1}   & \textbf{B-1}   \\\midrule
T5-base                          & 220M          & 38.40          & 81.02           & 52.64              & 41.28              & 56.53            \\
T5-large                         & 770M          & 40.77          & \textbf{90.62} & 57.04          & 43.82          & \textbf{59.59} \\\midrule
ClarET                           & 400M          & \textbf{41.04} & 87.18          & \textbf{57.47} & \textbf{48.75} & 58.88          \\\bottomrule
\end{tabular}
\caption{\small Fine-tuning generation results to compare with larger pre-trained models. Column names are datasets corresponding to those of Table~\ref{tab:gen_result}. See Appendix~\ref{app:sec:full_t5_res} for full results of T5.}
\label{tab:large_model}
\end{table}

\paragraph{Comparison with Larger Model.} 
A trend of pre-training models follows the law of `larger models for better performance' but a crucial research question is `how to perform competitively with fewer computation resources'.
To answer, we show extra fine-tuning results on the five generation datasets in Table~\ref{tab:large_model} to compare our ClarET (400M parameters) with T5-large (770M) and T5-base (220M). 
It is observed (i) with 3$\times$ scale, T5-large notably outperforms T5-base to support the above law and (ii) with almost half model size, our ClarET performs very competitively to T5-large (even better on 3 out of 5 tasks), verifying the significance of our objectives towards event-related knowledge.

\paragraph{Difficulty of Event Generation.}

To exhibit the \textit{learning difficulty} in pre-training (as stated in \S\ref{sec:whole_event_recover}) and the effectiveness of our novel learning objectives, we conduct another ablation setting in Table~\ref{tab:Event_ppl}. 
It is observed that ClarET achieves better event-level perplexity (ePPL), verifying the two novel objectives promote event generations and reduce difficulty of decoding.

\begin{figure}[t]
  \centering
  \includegraphics[width=0.98\linewidth]{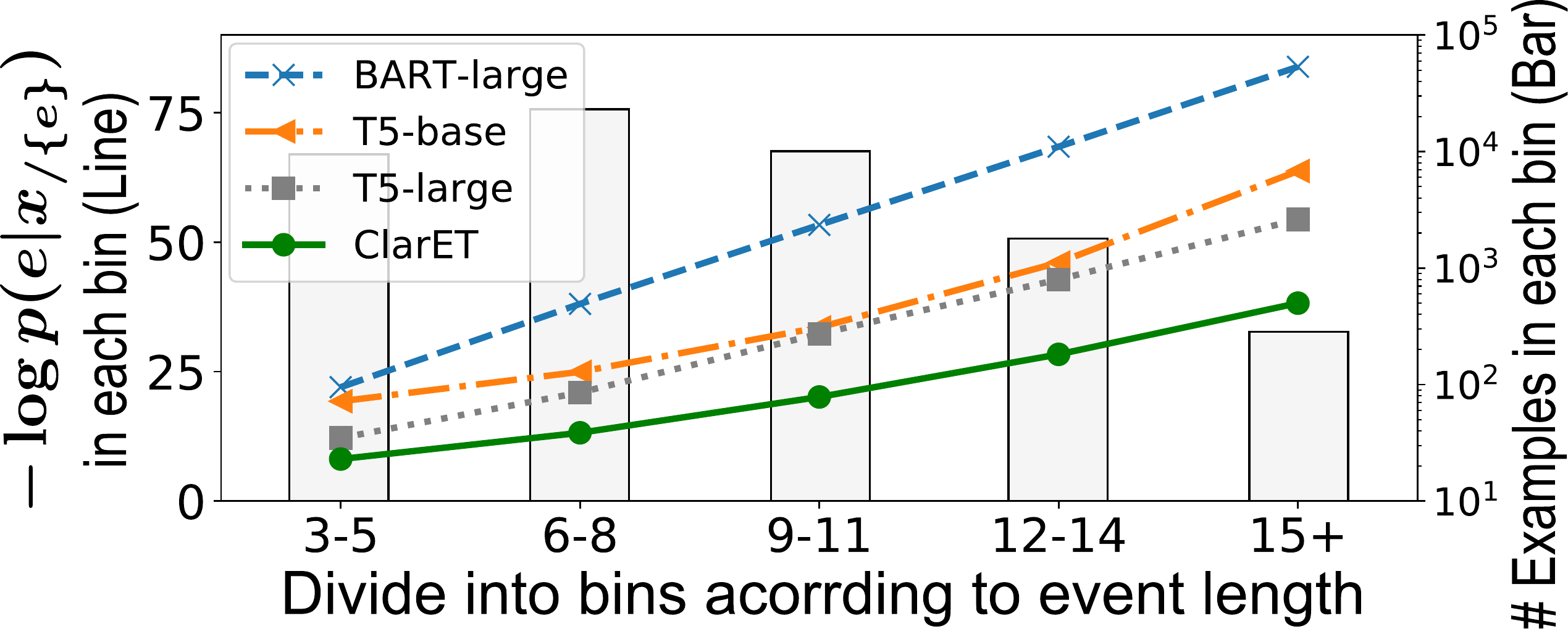}
  \caption{\small Event generation performance on the event-masked dev set (refer to Table~\ref{tab:Event_ppl}) with event-length bins. 
}\label{fig:long_span_gen}
\end{figure}
\paragraph{Long-span Event Generation.} 
To further check if ClarET is more competitive on longer-span event generation, we compare it with BART-large and T5-base/-large by `$-\log$' of Eq.(\ref{equ:obj_cag_base}). 
Different from recovering paradigm of others, we follow the denoising paradigm to implement BART and calculate its score by considering the masked part in decoding. 
Figure~\ref{fig:long_span_gen} shows that 
\textit{(1) Line Chart:} the gap between ClarET and the others becomes larger with event length increasing as the general-purpose models only consider short-span masking in pre-training, leading to inferior event generation;
and \textit{(2) Bar Chart:} as for data distribution, although a majority of data falls into the 6-8 bin, there are still many examples with event length greater than nine.

\begin{figure}[t]
\centering
\includegraphics[width=0.9\linewidth]{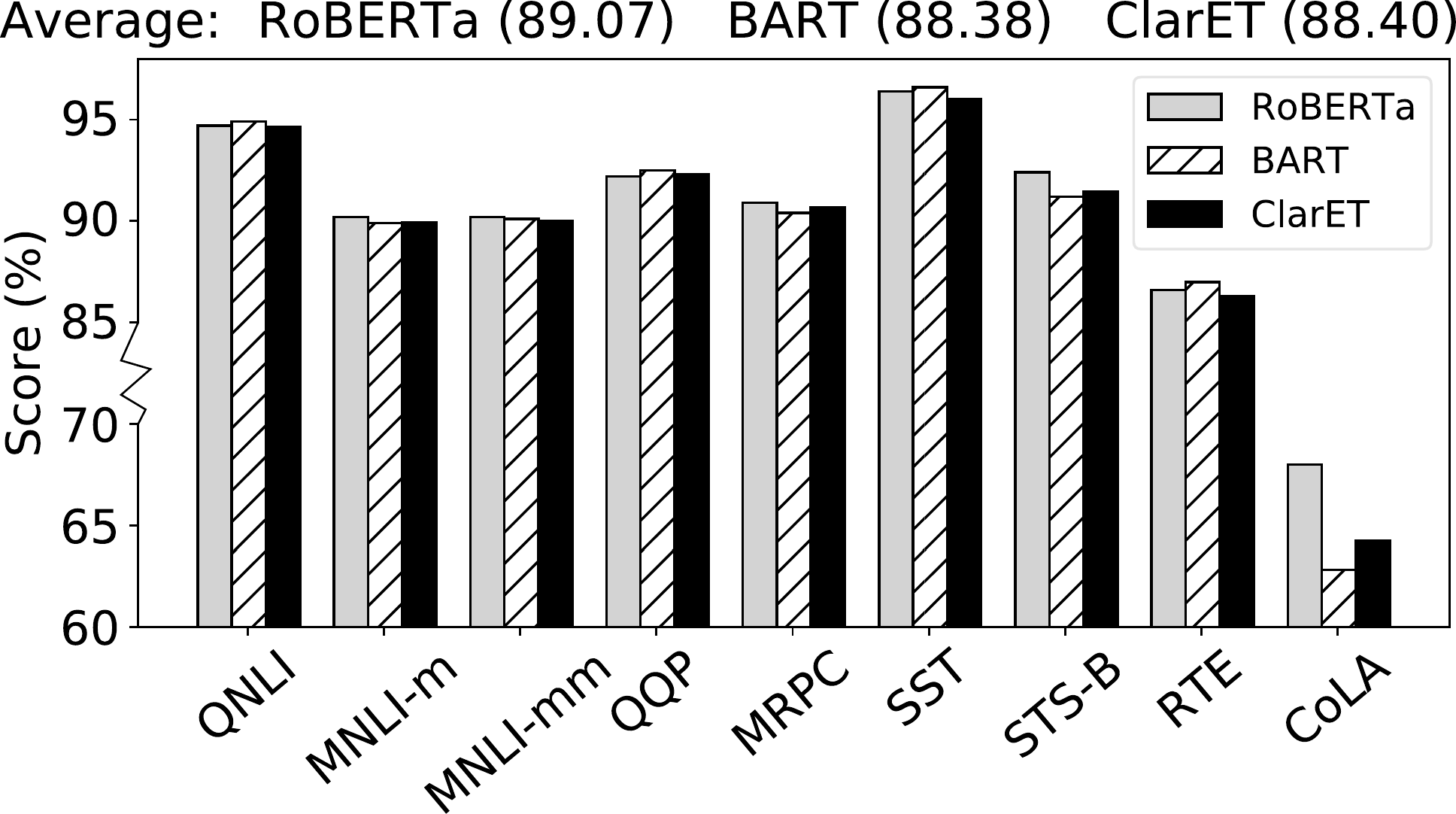}
\caption{\small Fine-tuning results on GLUE dev, which verifies ClarET retains BART's natural language understanding ability. 
}
\label{fig:glue_result}
\end{figure}
\paragraph{Natural Language Understanding (NLU).}
Our basic model, BART-large, is presented for general NLU tasks. To exhibit our minor event-centric continual pre-training would not interfere its NLU ability, we conduct fine-tuning experiments on GLUE benchmark \cite{Wang19GLUE} as in Figure~\ref{fig:glue_result}. 
It is observed that, although slightly surpassed by the discriminative RoBERTa model, fine-tuning BART and ClarET achieve very comparable results, which verifies ClarET's retention of NLU capability.

\subsection{Case Study and Error Analysis}

\begin{figure}[t]
    \centering
    \includegraphics[width=1.0\linewidth]{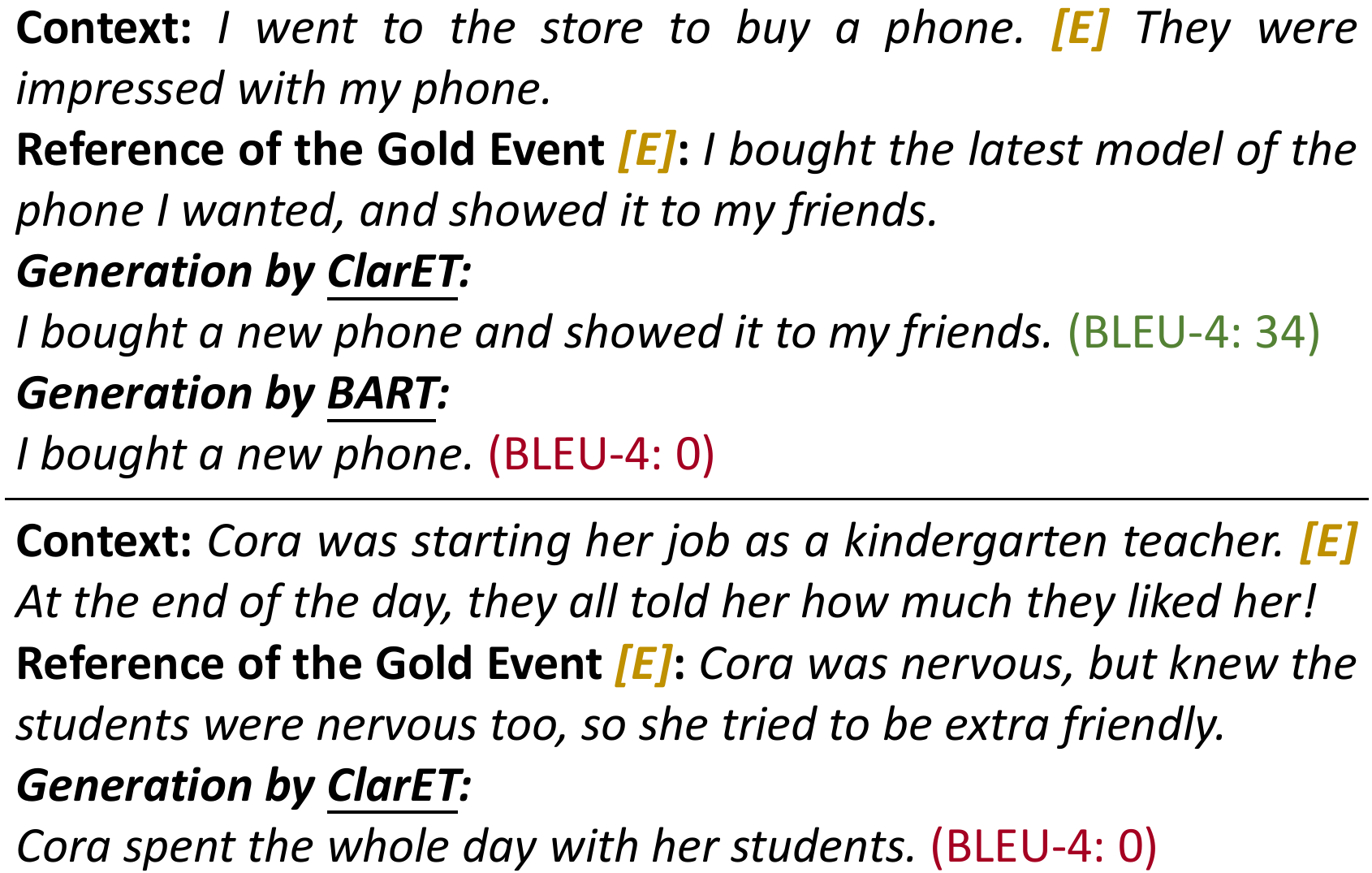}
    \caption{\small Case study \& error analysis on abductive reasoning. 
    }
    \label{fig:case_study}
\end{figure}

\paragraph{Case Study.}

As the first case in Figure~\ref{fig:case_study}, we conduct a case study on generative abductive reasoning task, where the fine-tuned ClarET generates an event semantically close to the gold reference, but the BART does not. 
BART only generates a part of the answer but ignores the event-correlations from `\textit{They were impressed with my phone}', while ClarET completely captures the correlations in the contexts (e.g., `\textit{to buy a phone}' and `\textit{They were impressed}',) and generate a much better result.

\begin{figure}[t]
    \centering
    \includegraphics[width=1.0\linewidth]{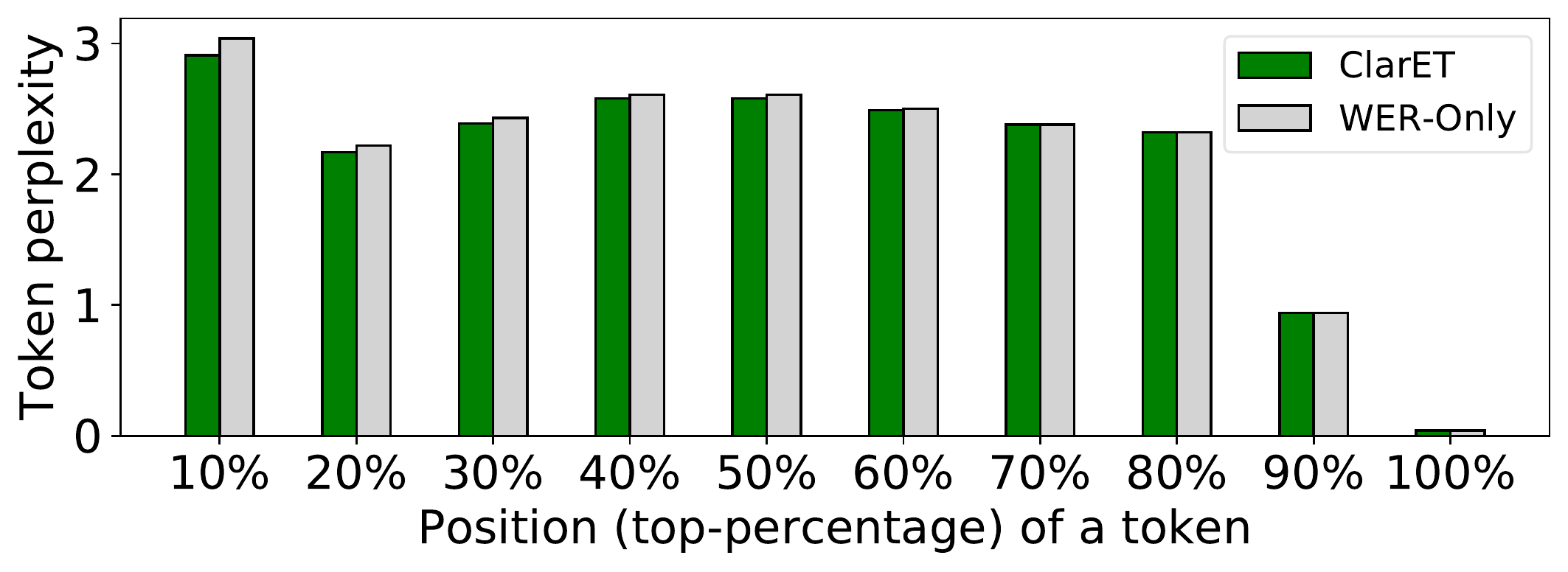}
    \caption{\small Token-level perplexity w.r.t tokens' percentage positions in events on held-out dev set. 
    }
    \label{fig:position-ppl}
\end{figure}

\paragraph{Error Analysis and Limitation.}
The second case in Figure~\ref{fig:case_study} shows that our ClarET is ineffective when the gold event is very complicated. 
In detail, the model focus only on `\textit{at the end of the day}' to generate `\textit{... spent the whole day ...}' but ignore very subtle contexts, e.g., `\textit{starting her job ... teacher}' and `\textit{they liked her}'. 
To expand, we found a problem in long-event decoding by pilot experiments. 
As shown in Figure~\ref{fig:position-ppl}, it is observed that 
the gap of token-level perplexity between ClarET and WER-only gradually diminishes. This is because the subsequent tokens in an event can be generated on the basis of previous generations on the decoder side, rather than context-aware representations from the encoder side.
While a long span is masked, the model can see previous tokens in an event (i.e., $e_{<t}$) in decoding and incline to perform the $t$-th prediction based on $e_{<t}$ but not $x_{/\{e\}}$, especially with a larger $t$. As a result, the model would `cheat' in the generation but learn decoder-side language modeling rather than context-aware representations. In the future, we will exploit this problem. 
Besides, due to computation resources, we choose the model size with 400M and continual pre-training in 90h, limiting the performance.

\section{Conclusion}
We present a novel correlation-aware context-to-event Transformer to self-supervisedly learn event-correlation knowledge from text corpus and benefit various event-centric reasoning scenarios. 
Besides SoTA fine-tuning results on 5 generation and 4 classification tasks, we conduct zero-/few-shot learning and extensive ablation studies to exhibit our model's effectiveness. 
Lastly, we find our model is competitive to a twice larger general-purpose model, reduces learning difficulty for event generation, and retains NLU ability from its basic model. Although this work learns context-to-event knowledge, our self-supervised objectives are applicable to other semantically-meaningful text units besides events. For example, text units can be entities and concepts to learn relational and commonsense knowledge, which can benefit more downstream tasks.

\section{Ethical Statement}
This work does not involve any sensitive data, but only public unlabeled corpora, i.e., BookCorpus \cite{Zhu15Aligning} pre-processed by \citet{Zhou21EventBERT}, and crowd-sourced datasets released in previous works, including \emph{\dataset{}} \cite{Bhagavatula20Abductive}, TIMETRAVEL \cite{Qin19Counterfactual}, APSI \cite{Zhang20Analogous}, MCNC \cite{Li18Constructing}, ROCStories \cite{mori2020finding}.

\bibliographystyle{acl_natbib}
\bibliography{ref}

\appendix

\section{Examples from Mined Pre-training Corpus} \label{app:sec:pre-train-example}
There are some mined pre-training examples shown in Table~\ref{tab:pretrain_example}. As in \citep{Zhou21EventBERT}, an example includes a paragraph, events, a selected positive event, connectives of the positive event, and sampled negative events of the positive event. 

\begin{table*}[t]\small
\centering
\begin{tabular}{ll}
\toprule
\multicolumn{2}{c}{\textbf{Example 1}}    \\
\midrule
\textbf{Paragraph}                        & \begin{tabular}[c]{@{}l@{}} It was only months later, when she saw her friend's thin gaunt face, her swollen belly and her quiet \\ desperation, that she had come to her senses. Then she had been filled with a combination of burning\\ rage and deep shame. This had endured over the years undiminished.\end{tabular} \\ \midrule
\textbf{Positive Event}                   & she had been filled with a   combination of burning rage     \\ \midrule
\textbf{Connectives}      & when; then    \\\midrule
\multirow{7}{*}{\textbf{Negative Events}} & he had been loaded with a lot of   vampire venom             \\
                                          & she had   been trained in the art of gentler speech          \\
                                          & I had   been blessed with some sort of fire ability          \\
                                          & he had   been transformed into a piece of living statuary    \\
                                          & he had   beened from a block of pale marble           \\
                                          & I had   been circumci sculptsed in the age of infantile apathy      \\
                                          & ...       \\\midrule\midrule
\multicolumn{2}{c}{\textbf{Example 2}}    \\
\midrule
\textbf{Paragraph}                        & \begin{tabular}[c]{@{}l@{}}Then,   when she turned twenty one at the end of last year, she had decided to act on   it. A driver's \\ license
was something she needed for her business and the identity papers which went with it were \\ needed for a range of other reasons,   such as enrolling Catherine for school at the start of this year.\end{tabular}
\\\midrule
\textbf{Positive Event}                   & papers which went with it were   needed for a range of other reasons               \\\midrule
\textbf{Connectives}      & then;  when; and  \\\midrule
\multirow{7}{*}{\textbf{Negative Events}} & bookcases that stood against it   had opened like a pair of French doors           \\
                                          & it had a   little bit of magic in it , just for Lizzie                             \\
                                          & it only   gave her a place for a couple of days                                    \\
                                          & which   occasionally crossed a small ridge sometimes of gravel , sometimes of sand \\
                                          & that she   was going shopping in the city with a couple of other girls             \\
                                          & publish   that proposal in the paper for three weeks                              \\ 
                                          & ...                     \\\bottomrule
\end{tabular}%
\caption{\small Some mined pre-training examples.}
\label{tab:pretrain_example}
\end{table*}

\begin{table*}[t]\small
\centering
\begin{tabular}{lcccccccccccc}
\toprule
\textbf{} & \multicolumn{3}{c}{\textbf{\begin{tabular}[c]{@{}c@{}}Abductive C.S. \\ Reasoning\end{tabular}}} & \multicolumn{3}{c}{\textbf{\begin{tabular}[c]{@{}c@{}}Counterfactual \\    Story\end{tabular}}} & \multicolumn{2}{c}{\textbf{\begin{tabular}[c]{@{}c@{}}Story Ending \\ Generation\end{tabular}}} & \multicolumn{2}{c}{\textbf{\begin{tabular}[c]{@{}c@{}}C.S. Story \\ Generation\end{tabular}}} & \multicolumn{2}{c}{\textbf{\begin{tabular}[c]{@{}c@{}}Event Process \\ Completion\end{tabular}}} \\ \cmidrule(lr){2-4} \cmidrule(lr){5-7} \cmidrule(lr){8-9} \cmidrule(lr){10-11} \cmidrule(lr){12-13}
          & \textbf{B-4}                   & \textbf{R-L}                   & \textbf{BERT}                  & \textbf{B-4}                   & \textbf{R-L}                  & \textbf{BERT}                  & \textbf{B-1}                                   & \textbf{B-2}                                   & \textbf{B-1}                                  & \textbf{B-2}                                  & \textbf{B-1}                                    & \textbf{B-2}                                   \\\midrule
T5-base   & 15.65                         & 38.40                         & 55.98                         & 81.02                        & 75.95                        & 79.12                       & 52.64                                         & 17.55                                        & 41.28                                        & 13.76                                       & 56.53                                         & 18.84                                         \\\midrule
T5-large  & 17.75                       & 40.77                       & 57.20                       & 90.62                       & 84.03                       & 83.14                      & 57.04                                      & 18.45                                     & 43.82                                     & 14.61                                    & 59.59                                      & 19.86      \\\bottomrule
\end{tabular}
\caption{Full results of T5-base and T5-large on generation tasks.}
\label{tab:t5_results}
\end{table*}

\begin{table*}[t]\small
\centering
\begin{tabular}{ll}
\toprule
\multicolumn{2}{c}{\textbf{Dataset 1: \emph{\dataset{}} ($\alpha$NLG)}}                                                                                                                                                                                                                                                                                                                                                                                 \\\midrule
\textbf{Input} & \begin{tabular}[c]{@{}l@{}}Observation1: The hayride was in   October.     \\ Observation2: It was the   perfect start to the fall season.\end{tabular}                                                                                                                                                                                                                                         \\\midrule
\textbf{Label} & Keeping tradition we drank hot   cocoa on the ride.                                                                                                                                                                                                                                                                                                                                             \\\midrule\midrule
\multicolumn{2}{c}{\textbf{Dataset 2: TIMETRAVEL}}                                                                                                                                                                                                                                                                                                                                                                          \\\midrule
\textbf{Input} & \begin{tabular}[c]{@{}l@{}}Premise: On my way to work I   stopped to get some coffee.    \\ Initial:   I went through the drive through and placed my order.     \\ Original\_ending: I paid the cashier and   patiently waited for my drink. \\ When she handed me the drink, the lid came off   and spilled on me. \\ The coffee hurt and I had to go home and change   clothes. \\ Counterfactual: I went inside to place my order. \end{tabular} \\\midrule
\textbf{Label} & \begin{tabular}[c]{@{}l@{}}I paid the cashier and patiently   waited at the counter for my drink. \\ When she handed me the drink, the lid   came off and spilled on me. \\ The coffee hurt and I had to go home and change   clothes.\end{tabular}                                                                                                                                             \\\midrule\midrule
\multicolumn{2}{c}{\textbf{Dataset 3: Story Ending   Generation}}                                                                                                                                                                                                                                                                                                                                                           \\\midrule
\textbf{Input} & \begin{tabular}[c]{@{}l@{}}Dan's parents were overweight.   \\ Dan was overweight as well. \\ The doctors told his parents it was unhealthy.   \\ His parents understood and decided to make a change.\end{tabular}                                                                                                                                                                             \\\midrule
\textbf{Label} & They got themselves and Dan on a diet.                                                                                                                                                                                                                                                                                                                                                          \\\midrule\midrule
\multicolumn{2}{c}{\textbf{Dataset 4: Commonsense Story   Generation}}                                                                                                                                                                                                                                                                                                                                                      \\\midrule
\textbf{Input} & Carrie had just learned how to ride a bike.                                                                                                                                                                                                                                                                                                                                                     \\\midrule
\textbf{Label} & \begin{tabular}[c]{@{}l@{}}She didn't have a bike of her   own.\\ Carrie would sneak rides on her sister's bike. \\ She got nervous on a hill   and crashed into a wall. \\ The bike frame bent and Carrie got a deep gash on   her leg.\end{tabular}                                                                                                                                           \\\midrule\midrule
\multicolumn{2}{c}{\textbf{Dataset 5: APSI}}                                                                                                                                                                                                                                                                                                                                                                                \\\midrule
\textbf{Input} & Process name: Treat Pain.   Process: Identify cause. learn injury.                                                                                                                                                                                                                                                                                                                               \\\midrule
\textbf{Label} & Recognize symptom.                                                                                                                                                                                                                                                                                                                                                                              \\\midrule\midrule
\multicolumn{2}{c}{\textbf{Dataset 6: Multi-choice   narrative cloze (MCNC)}}                                                                                                                                                                                                                                                                                                                                               \\\midrule
\textbf{Input} & \begin{tabular}[c]{@{}l@{}}Context: compare basketball.   buck get basketball. whirl basketball bench. shout out basketball   center.     \\ Options: A. look   basketball. B. weaken basketball. C. throw basketball lot youngster. \\ D.   client deny basketball. E. client deny basketball.\end{tabular}                                                                              \\\midrule
\textbf{Label} & C                                                                                                                                                                                                                                                                                                                                                                                               \\\midrule\midrule
\multicolumn{2}{c}{\textbf{Dataset 7: \emph{\dataset{}} ($\alpha$NLI)}}                                                                                                                                                                                                                                                                                                                                                                                 \\\midrule
\textbf{Input} & \begin{tabular}[c]{@{}l@{}}Observation1: Chad went to get   the wheel alignment measured on his car.      \\ Observation2: The mechanic provided a working alignment with new body   work.    \\ Hypothesis1: Chad was waiting   for his car to be washed.      \\ Hypothesis2: Chad was waiting for his car to be finished.\end{tabular}                                                       \\\midrule
\textbf{Label} & 2                                                                                                                                                                                                                                                                                                                                                                                               \\\midrule\midrule
\multicolumn{2}{c}{\textbf{Dataset 8: ROCStories}}                                                                                                                                                                                                                                                                                                                                                                          \\\midrule
\textbf{Input} & \begin{tabular}[c]{@{}l@{}}Laverne needs to prepare   something for her friend's party. \\ She decides to bake a batch of brownies.   \\ Laverne tests one of the brownies to make sure it is delicious. \\ The brownies   are so delicious Laverne eats two of them.\end{tabular}                                                                                                              \\\midrule
\textbf{Label} & 3                                                                                                                                                                                                                                                                                                                                                                                               \\\midrule\midrule
\multicolumn{2}{c}{\textbf{Dataset 9: Story Cloze Test}}                                                                                                                                                                                                                                                                                                                                                                    \\\midrule
\textbf{Input} & \begin{tabular}[c]{@{}l@{}}Context: Rick grew up in a   troubled household. He never found good support in family, and turned to   gangs. \\ It wasn't long before Rick got shot in a robbery. The incident caused   him to turn a new leaf.     \\ Options: A.   He is happy now. B. He joined a gang.\end{tabular}                                                                            \\\midrule
\textbf{Label} & A                                  \\\bottomrule  
\end{tabular}
\caption{The training examples on different datasets.}
\label{tab:train_example}
\end{table*}

\section{More Details} \label{app:sec:more_details}

\subsection{BART Pre-training Resources} \label{app:sec:bart_pretrain_resource}

In this section, we analyze BART pre-training resources in terms of text corpora and computation resources. 

As for tokens in BART pre-training corpora, BART paper \citep{Lewis20BART} claims using the same corpora as in RoBERTa \citep{Liu19RoBERTa} and T5 paper \citep{Raffel20Exploring} states RoBERTa uses a 2.2T-token text corpus. Thus, we adopt `2.2T' as the number in the main paper.

As for BART pre-training computation overheads, the contributor of BART official code repository said `\textit{We trained for around 11-12 days on 256 gpus.}' at \url{https://github.com/pytorch/fairseq/issues/1525}, so the BART pre-training takes from 67584 to 73728 GPU hours.
Thus, we use `70,000' as the number in the main paper. 

\subsection{Full Results of T5 Model} \label{app:sec:full_t5_res}

The full results of T5-base and T5-large on the five generation tasks are shown in Table~\ref{tab:t5_results}.

\subsection{Connectives in Paragraph} \label{app:sec:connectives}

As stated by \citet{Zhou21EventBERT}, connectives (i.e., discourse relations in the contexts) play important roles to express correlations among events. Therefore, we also find every possible connective $r$ to each $(x,e)$ where $r$ is a connective in $x$, which immediately links to the verb of $e$ on the parsing tree of $x$. To leverage the connectives, we also apply the correct event selection in prompt-based event locating objective to $r$ its negatives $\{\bar r\}_1^M$, as correct connective selection. Here, $\bar r$ is randomly sampled from discourse relations in the PDTB annotation manual \citep{webber2019penn}. 
At 20\% times, we use correct connective selection to replace correct event selection in the prompt-based event locating objective.

\section{Details of Evaluation Datasets} \label{app:sec:dataset}
We detail the nine evaluation datasets in the following. The training example in each dataset is shown in Table \ref{tab:train_example}.
\begin{itemize}
    \item \textbf{\emph{\dataset{}} ($\alpha$NLG).} Given two observations in natural language, it aims to generate an explicative hypothesis between them. We follow the official data split \cite{Bhagavatula20Abductive} with 169,654/1,532/3,059 in training/dev/test. 
    
    \item \textbf{TIMETRAVEL.} Given an original story and a counterfactual event, it aims to rewrite the subsequent events to complete a story, which is compatible with the counterfactual event. We follow the official data split \cite{Qin19Counterfactual} with 98,159/5,613/7,484 in training/dev/test.
    
    \item \textbf{Story Ending Generation.} We evaluate the story ending generation based on ROCStories, which aims to generate a story ending for a given story context. We follow the data split \cite{Guan19Story} with 90,000/4,081/4,081 in training/dev/test.
    
    \item \textbf{Commonsense Story Generation.} It is based on ROCStories.
    Given a leading context, it aims to generate a reasonable story. We follow the data split \cite{Guan20Knowledge} with 88,344/4,908/4,909 in training/dev/test.
 
    \item \textbf{APSI.} We evaluate event process completion on the APSI dataset, where the goal is to generate a subevent for a given event context. We follow the data split \cite{Zhang20Analogous} with 13,501/1,316 in training/test.
    
    \item \textbf{Multi-choice narrative cloze (MCNC).} Given an event chain, it aims to predict the subsequent event from 5 candidates.
    We follow the data split \cite{Li18Constructing} with 140,331/10,000/10,000 in training/dev/test.
    
    \item \textbf{\emph{\dataset{}} ($\alpha$NLI).} Given two observations in natural language, it aims to choose the most explicative hypothesis from 2 candidates.
    We follow the data split \cite{Bhagavatula20Abductive} with 169,654/1532 samples in training/dev.
    
    \item \textbf{ROCStories.} We follow \cite{mori2020finding} to use ROCStories for narrative incoherence detection. A random sentence is removed for each five-sentence story, and the goal is to predict the missing position.
    We follow the data split \cite{mori2020finding} with 78,528/9,816/9,817 in training/dev/test.
    
    \item \textbf{Story Cloze Test.} Given a 4-sentence context, it aims to select the right ending from two alternative endings.
    We follow the data split \cite{Mostafazadeh16Corpus} with 98,161/1,871/1,871 in training/dev/test.
\end{itemize}

\section{Detailed Evaluation Results}
We detail the full results on nine evaluation datasets as follows. 

\subsection{Generation Tasks} \label{app:sec:gen_task}
Generation tasks include abductive commonsense reasoning ($\alpha$NLG), counterfactual story generation, 
story ending generation, 
commonsense story generation, 
and event process completion. 
The detailed results of these generation tasks are shown in Table \ref{tab:anlg}, Table \ref{tab:cs}, Table \ref{tab:seg}, Table \ref{tab:csg}, and Table \ref{tab:epc}, respectively. ClarET achieves state-of-the-art performance on all five generation tasks. In addition, pre-trained language models show their strong generation ability on story generation tasks, i.e., story ending generation and commonsense story generation.

\subsection{Classification Tasks} \label{app:sec:cls_task}
Classification tasks include script reasoning, 
abductive commonsense reasoning ($\alpha$NLI), 
narrative incoherence detection, 
and story cloze test.
The detailed results of these classification tasks are shown in Table \ref{tab:sr}, Table \ref{tab:anli}, Table \ref{tab:nid}, and Table \ref{tab:sct}, respectively. Compared with unified language models, ClarET achieves state-of-the-art performance. Although strong discriminative models show their great ability on classification tasks, ClarET still achieves competitive performance.

\begin{table}[t]\small
\centering
\resizebox{\linewidth}{!}{
\setlength\tabcolsep{2pt}
\begin{tabular}{lccc}
\toprule
\textbf{Method}   & \textbf{B-4} & \textbf{R-L} & \textbf{BERT}  \\ \midrule
GPT2-Fixed \cite{Bhagavatula20Abductive}       & 0.00            & 9.99             & 36.69          \\
O1-O2-Only \cite{Bhagavatula20Abductive}       & 2.23            & 22.83            & 48.74          \\
COMeT-T+GPT2 \cite{Bhagavatula20Abductive}   & 2.29            & 22.51            & 48.46          \\
COMeT-E+GPT2 \cite{Bhagavatula20Abductive}   & 3.03         \textit{}   & 22.93            & 48.52          \\
Fine-tuned GPT2-L \cite{Dong21On-the-Fly} & 13.52           & 18.01            & -              \\
GRF \cite{Ji20Language}              & 11.62           & 34.62            & -              \\
BART-large \cite{Lewis20BART}       & 16.47           & 38.73            & 56.36          \\
GLM-large \cite{Du21All}        & 7.79            & 25.54            & 54.85          \\
ClarET         & \textbf{17.67}  & \textbf{41.04}   & \textbf{57.31} \\ \bottomrule
\end{tabular}}
\caption{\small Results on the Abductive Commonsense Reasoning ($\alpha$NLG).
}
\label{tab:anlg}
\end{table}

\begin{table}[t]\small
\centering
\setlength\tabcolsep{2pt}
\begin{tabular}{lccc}
\toprule
\textbf{Method}     & \textbf{B-4} & \textbf{R-L} & \textbf{BERT}  \\ \midrule
Human \cite{Qin19Counterfactual}   & 64.93          & 67.64          & 61.87             \\
GPT2-S \cite{radford2019language}        & 69.27 & 65.72 & 60.53 \\
GPT2-M+Rec+CF \cite{Qin20Back} & 75.92          & 70.93          & 62.49             \\
GPT2-M+Sup \cite{Qin20Back}       & 75.71          & 72.72          & 62.39             \\
BART-large \cite{Lewis20BART}    & 82.91 & 76.44 & 79.50 \\
GLM-large \cite{Du21All}     & 75.81 & 70.03 & 68.23 \\
ClarET           & \textbf{87.18} & \textbf{80.74} & \textbf{81.48}    \\ \bottomrule
\end{tabular}
\caption{\small Results on the Counterfactual Story.
}
\label{tab:cs}
\end{table}

\begin{table}[t]\small
\centering
\begin{tabular}{lcc}
\toprule
\textbf{Method}      & \textbf{B-1} & \textbf{B-2} \\ \midrule
Seq2Seq \cite{Luong15Effective}     & 18.50           & 5.90            \\
Transformer \cite{Vaswani17Attention} & 17.40           & 6.00            \\
GCN \cite{Yang18Graph}         & 17.60           & 6.20            \\
IE+MSA \cite{Guan19Story}      & 24.40           & 7.80            \\
T-CVAE \cite{Wang19T-CVAE}     & 24.30           & 7.70            \\
Plan\&Write \cite{Yao19Plan-and-Write} & 24.40           & 8.40            \\
MGCN-DP \cite{Huang21Story}    & 24.60           & 8.60            \\
GPT2-S \cite{radford2019language} & 39.23           & 13.08           \\
BART-large \cite{Lewis20BART}  & 54.22           & 18.07           \\
ClarET   & \textbf{57.47}  & \textbf{19.16}  \\ \bottomrule
\end{tabular}
\caption{\small Results on the Story Ending Generation.
}
\label{tab:seg}
\end{table}

\begin{table}[t]\small
\centering
\begin{tabular}{lcc}
\toprule
\textbf{Method} & \textbf{B-1} & \textbf{B-2} \\ \midrule
ConvS2S \cite{Gehring17Convolutional}        & 31.20           & 13.20           \\
Fusion \cite{Fan18Hierarchical}         & 32.20           & 13.70           \\
Plan\&Write \cite{Yao19Plan-and-Write}    & 30.80           & 12.60           \\
SKRL \cite{Xu18Skeleton}           & 26.70           & 8.80            \\
DSRL \cite{Fan19Strategies}           & 29.30           & 11.70           \\
GPT2-S \cite{Guan20Knowledge}    & 32.20           & 14.10           \\
KE-GPT2 \cite{Guan20Knowledge}        & 32.60           & 14.30           \\
BART-large \cite{Lewis20BART}     & 45.24           & 15.08           \\
ClarET       & \textbf{48.75}  & \textbf{16.25}  \\ \bottomrule
\end{tabular}
\caption{\small Results on the Commonsense Story Generation.
}
\label{tab:csg}
\end{table}

\begin{table}[t]\small
\centering
\begin{tabular}{lcc}
\toprule
\textbf{Method} & \textbf{B-1} & \textbf{B-2} \\ \midrule
GPT2-S \cite{radford2019language}    & 35.25           & 11.75           \\
GPT2-M \cite{radford2019language}    & 45.43 & 14.81 \\
BART-large \cite{Lewis20BART}     & 56.25           & 18.75           \\
GLM-large \cite{Du21All}  & 57.34 & 19.11 \\
ClarET       & \textbf{58.88}  & \textbf{19.74}  \\ \bottomrule
\end{tabular}
\caption{\small Results on the Event Process Completion.
}
\label{tab:epc}
\end{table}

\begin{table}[t]\small
\centering
\setlength\tabcolsep{2pt}
\begin{tabular}{lc}
\toprule
\textbf{Method} & \textbf{ACC} \\
\hline\hline
\multicolumn{2}{l}{\textit{Discriminative Model}}  \\ \hline
Random                          & 20.00    \\
Event-Comp \cite{Granroth-Wilding16What}                     & 49.57    \\
PairLSTM \cite{Wang17Integrating}                       & 50.83    \\
SGNN \cite{Li18Constructing}                           & 52.45    \\
SGNN + Int\&Senti \cite{Ding19Event}              & 56.03    \\
RoBERTa-base \cite{Liu19RoBERTa}                   & 56.23    \\
RoBERTa-large \cite{Liu19RoBERTa}                  & 61.53    \\
RoBERTa + Rep. Fusion \cite{Lv20Integrating} & 58.66    \\
EventBERT \cite{Zhou21EventBERT}                      & 63.50    \\
RoBERTa + Kown. Model \cite{Zhou21Modeling}    & 63.62    \\
\hline\hline
\multicolumn{2}{l}{\textit{Unified Model}}  \\ \hline
BART-large \cite{Lewis20BART}                     & 61.34    \\
ClarET                       & \textbf{64.61}    \\ \bottomrule
\end{tabular}
\caption{\small Results on the Script Reasoning.
}
\label{tab:sr}
\end{table}

\begin{table}[t]\small
\centering
\begin{tabular}{lc}
\toprule
\textbf{Method} & \textbf{ACC} \\ 
\hline\hline
\multicolumn{2}{l}{\textit{Discriminative Model}}  \\ \hline
Random          & 50.00    \\
BERT-base \cite{Devlin19BERT}      & 61.88    \\
ERNIE \cite{Zhang19ERNIE}          & 63.04    \\
KnowBERT \cite{Peters19Knowledge}       & 63.18    \\
BERT-large \cite{Devlin19BERT}     & 66.75    \\
RoBERTa-large \cite{Liu19RoBERTa}   & 82.35    \\
EventBERT \cite{Zhou21EventBERT}       & \textbf{85.51}    \\
\hline\hline
\multicolumn{2}{l}{\textit{Unified Model}}  \\ \hline
T5-base \cite{Raffel20Exploring}        & 61.10    \\
T5-large \cite{Raffel20Exploring}       & 77.80    \\
BART-large \cite{Lewis20BART}     & 80.74    \\
GLM-large \cite{Du21All}      & 65.27    \\
CALM-large \cite{Zhou21Pre-training}     & 77.12    \\
UNICORN \cite{Lourie21UNICORN}        & 79.50    \\
ClarET       & 82.77    \\ \bottomrule
\end{tabular}
\caption{\small Results on the Abductive Commonsense  Reasoning ($\alpha$NLI).
}
\label{tab:anli}
\end{table}

\begin{table}[t]\small
\centering
\begin{tabular}{lc}
\toprule
\textbf{Method} & \textbf{ACC} \\
\hline\hline
\multicolumn{2}{l}{\textit{Discriminative Model}}  \\ \hline
Random           & 20.00    \\
RoBERTa-large \cite{Liu19RoBERTa}   & 73.94    \\
Max-pool Context \cite{mori2020finding} & 35.00    \\
GRU Context \cite{mori2020finding}     & 52.20    \\
EventBERT \cite{Zhou21EventBERT}       & \textbf{75.03}    \\
\hline\hline
\multicolumn{2}{l}{\textit{Unified Model}}  \\ \hline
BART-large \cite{Lewis20BART}      & 72.48    \\
ClarET        & 74.88    \\ \bottomrule
\end{tabular}
\caption{\small Results on the Narrative Incoherence Detection.}
\label{tab:nid}
\end{table}

\begin{table}[t]\small
\centering
\setlength\tabcolsep{2pt}
\begin{tabular}{lc}
\toprule
\textbf{Method} & \textbf{ACC} \\
\hline\hline
\multicolumn{2}{l}{\textit{Discriminative Model}}  \\ \hline
Random                   & 50.00          \\
Hidden Coherence Model \cite{Chaturvedi17Story}  & 77.60          \\
val-LS-skip \cite{Srinivasan18Simple}             & 76.50          \\
RoBERTa-large \cite{Liu19RoBERTa}           & 87.10          \\
EventBERT \cite{Zhou21EventBERT}               & \textbf{91.33} \\
\hline\hline
\multicolumn{2}{l}{\textit{Unified Model}}  \\ \hline
Finetuned Transformer LM \cite{radford2018improving} & 86.50          \\
BART-large \cite{Lewis20BART}              & 87.01          \\
ClarET                & 91.18          \\ \bottomrule
\end{tabular}
\caption{\small Results on the Story Cloze Test.}
\label{tab:sct}
\end{table}

\end{document}